\documentclass{article}
\usepackage{ijcai24}

\usepackage{times}
\usepackage{soul}
\usepackage{url}
\usepackage[hidelinks]{hyperref}
\usepackage[utf8]{inputenc}
\usepackage[small]{caption}
\usepackage{graphicx}
\usepackage{amsmath}
\usepackage{amsthm}
\usepackage{booktabs}
\usepackage{algorithm}
\usepackage{algorithmic}
\usepackage[switch]{lineno}
\usepackage{amssymb}
\usepackage{bbm, dsfont}
\usepackage{caption}
\usepackage{float} 
\usepackage{subcaption}
\usepackage{ragged2e} 
\usepackage{makecell, multirow, tabularx}
\usepackage{diagbox}
\usepackage[table]{xcolor}
\usepackage{array}
\usepackage{booktabs}
\usepackage{float}

\title{Joint Input and Output Coordination for Class-Incremental Learning}

\author{
Shuai Wang$^{1,2}$
\and
Yibing Zhan$^3$\and
Yong Luo$^{1,2\ast}$\and
Han Hu$^4$\and
Wei Yu$^1$\footnote{Corresponding authors: Yong Luo, Wei Yu.}\and \\
Yonggang Wen$^5$\and
Dacheng Tao$^5$\\
\affiliations
$^1$Institute of Artificial Intelligence, School of Computer Science, Wuhan University, China.\\
$^2$ Hubei Luojia Laboratory, Wuhan, China.
$^3$JD Explore Academy, JD.com, Inc., China.\\
$^4$School of Information and Electronics, Beijing Institute of Technology, China.\\
$^5$College of Computing \& Data Science, Nanyang Technological University, Singapore.
\emails
wangshuai123@whu.edu.cn,
zhanyibing@jd.com,
luoyong@whu.edu.cn,
hhu@bit.edu.cn,
yuwei@whu.edu.cn,
ygwen@ntu.edu.sg,
dacheng.tao@ntu.edu.sg
}

\begin{document}

\maketitle

\begin{abstract}
Incremental learning is nontrivial due to severe catastrophic forgetting. Although storing a small amount of data on old tasks during incremental learning is a feasible solution, current strategies still do not 1) adequately address the class bias problem, and 2) alleviate the mutual interference between new and old tasks, and 3) consider the problem of class bias within tasks.
This motivates us to propose a joint input and output coordination (JIOC) mechanism to address these issues. This mechanism assigns different weights to different categories of data according to the gradient of the output score, and uses knowledge distillation (KD) to reduce the mutual interference between the outputs of old and new tasks. The proposed mechanism is general and flexible, and can be incorporated into different incremental learning approaches that use memory storage. Extensive experiments show that our mechanism can significantly improve their performance. 
\end{abstract}

\section{Introduction}
In recent years, 
incremental learning has attracted much attention since it can play an important role in a wide variety of fields, including unmanned driving~\cite{santoso2022data} and human-computer interaction~\cite{tschandl2020human}. Incremental learning is nontrivial since the parameters of deep models in the old tasks are often destroyed in the process of learning new tasks. This leads to the occurrence of catastrophic forgetting~\cite{french2002using}. How to well preserve past information and fully explore new knowledge has become a major challenge of incremental learning.

Existing incremental learning approaches mainly focus on memory storage replay~\cite{ahn2021ss,li2017learning,wu2019large,rebuffi2017icarl,yan2021dynamically}, model dynamic expansion~\cite{serra2018overcoming,mallya2018packnet}, and regularization constraints design~\cite{aljundi2019task}. Memory store replay has been demonstrated to be very effective, and it alleviates the destruction of old task weights by storing past data or simulating human memory. However, due to the privacy restriction and limited memory, the data to be accessed from old tasks are often quite scarce. This makes incremental learning models suffer from severe inter-task class bias, or known as the class imbalance issue between old and new tasks.

\begin{figure}[!t]
  \centering  \includegraphics[scale=0.33]{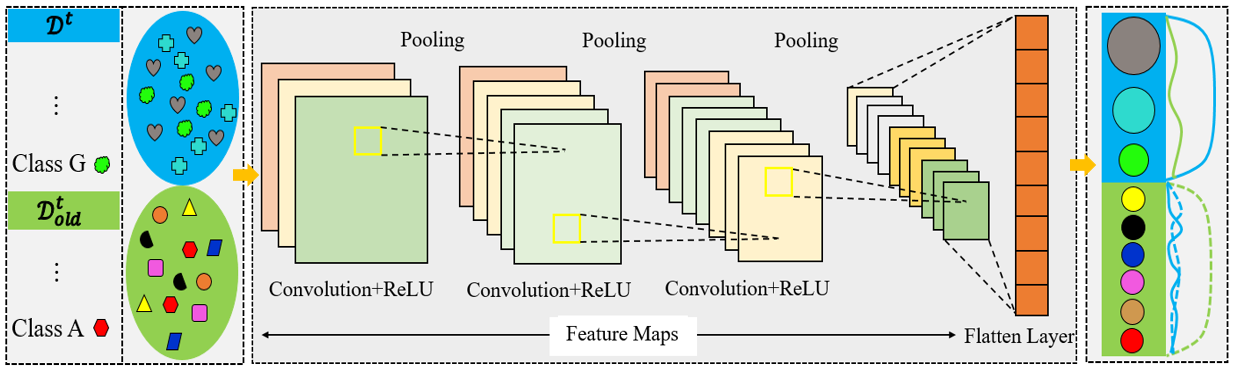}
  \caption{An illustration of the class imbalance and mutual interference issues. The difference in the number of input data for each class between tasks and within tasks makes the weights of fully connected layers greatly biased (neuron size). The output scores of data from old tasks ($1, \cdots, t-1$) on the classification heads of new task $t$ should approximate zero, but may be much larger than zero (green solid line) after training the new task model. The output scores of data from the new task on the classification heads of old tasks may be inconsistent before (blue dotted line) and after (blue solid line) updating the old task models.}
  \label{fig:class_bias}
\end{figure}

There exist some recent approaches~\cite{ahn2021ss,rebuffi2017icarl,yan2021dynamically} that alleviate the problem of class imbalance between old and new tasks by utilizing rescaling, balanced scoring, or softmax separating. 
Although these approaches can improve the performance to some extent, the problem of category imbalance still exists, since during the incremental learning progresses, the category imbalance becomes more severe as the number of sample categories continuously increase.
Moreover, the mutual interference between old and new tasks has not been well addressed. That is, only the predictions in old tasks are tried to be maintained, and the output scores of old task data on the classification heads of new tasks are not well suppressed. The output consistency of new task data on old classification heads before and after updating the new task model is also not considered. Besides, none of the existing approaches deal with the class bias within tasks.
An illustration is shown in Figure~\ref{fig:class_bias}.

In order to address these issues, we propose a joint input and output coordination (JIOC) mechanism, which enables incremental learning models to simultaneously alleviate the class imbalance and reduce the interference between the predictions of old and new tasks.
Specifically, different weights are adaptively assigned to different input data according to their gradients for the output scores during the training of the new task and updating of the old task models. Then the outputs of old task data on new classification heads are explicitly suppressed and knowledge distillation (KD)~\cite{menon2021statistical} is utilized for harmonization of the output scores based on the principle of human inductive memory~\cite{williams1999memory,redondo2011making}.

The main contributions are summarized as follows:
\begin{itemize}
    \item We propose a joint input and output coordination mechanism for incremental learning. As far as we are concerned, this is the first work that simultaneously adjusts input data and output layer for incremental learning;
    \item We design an adaptive input weighting strategy. The samples of different classes are weighted according to their gradients of the output scores. This alleviates the class bias problem both in and between tasks.
    \item We develop an output coordination strategy, which maintains the outputs of new task data on the old task classification heads before and after training, and suppresses the outputs of old task data on the new task classification heads.
\end{itemize}

The proposed method is general and flexible, and can be utilized as a plug-and-play tool for existing incremental learning approaches that use memory storage. To demonstrate the effectiveness of our mechanism, we incorporate it into some recent or competitive incremental learning approaches on multiple popular datasets (CIFAR10-LT, CIFAR100-LT, CIFAR100~\cite{krizhevsky2009learning}, MiniImagNet~\cite{vinyals2016matching}, TinyImageNet~\cite{le2015tiny} and Cub-200-2011~\cite{wah2011caltech}).
The results show that we can consistently improve the existing approaches, and the relative improvement is more than $10\%$ sometimes.

\section{Related Work}
\subsection{Incremental Learning}
Incremental learning~\cite{de2021continual} has received extensive attention in recent decades. In incremental learning, input data in new tasks are continuously used to extend the knowledge of existing models. This makes incremental learning manifest as a dynamic learning technique. An incremental learning model can be defined as one that meets the following conditions: (1) The model can learn useful knowledge from new task data; (2) The old task data that has been used to train the model does not need to be accessed or has a small amount of access; (3) It has a memory function for the knowledge that has been learned. The current study on incremental learning mainly focuses on domain incremental learning~\cite{mirza2022efficient,garg2022multi,mallya2018piggyback}, class-incremental learning~\cite{ahn2021ss,rebuffi2017icarl,yan2021dynamically,zhang2020class,liu2021adaptive}, and small sample incremental learning~\cite{tao2020few,cheraghian2021synthesized}.

There are many works on class-incremental learning (CIL), and most of these works overcome catastrophic forgetting by using knowledge distillation (KD) together with a small amount of old task data accessed. For example, DMC~\cite{zhang2020class} utilizes separate models for the new and old classes and trains the two models by combining double distillation. SPB~\cite{liu2021adaptive} utilizes cosine classifier and reciprocal adaptive weights, and a new method of learning class-independent knowledge and multi-view knowledge is designed to balance the stability-plasticity dilemma of incremental learning.

Although the above approaches can achieve promising performance sometimes, none of them address class bias within tasks, nor adequately address class bias between old and new tasks. Therefore, we propose joint input and output coordination (JIOC) mechanism that enables incremental learning models to alleviate class imbalance and reduce interference between the predictions of old and new tasks.

\subsection{Human Inductive Memory}
The inductive memory method is a unique ability of human beings. It causes the memorized content to be induced according to different attributes or categories; Subsequently, these contents are memorized by different categories or attributes. As early as 1999, Williams et al.~\cite{williams1999memory} investigated the relationship between memory for input and inductive learning of morphological rules relating to functional categories in a semi-artificial form of Italian. The ability to perform induction appears in the early age of human, while the underlying mechanisms remain unclear.
Therefore, Fisher et al.~\cite{fisher2005induction} demonstrated that category- and similarity-based induction should result in different memory traces and thus different memory accuracy.
Hayes et al.~\cite{Hayes2013} examined the development of the relationship between inductive reasoning and visual recognition memory, and demonstrated it through two studies. 
Inspired by human inductive memory, Geng et al.~\cite{geng2020dynamic} proposed a Dynamic Memory Induction Network (DMIN) to further address the small-sample challenge. These examples of inductive memory inspire us to propose an output distribution coordination mechanism.

\begin{figure*}[!t]
\centering
\includegraphics[scale=0.5]{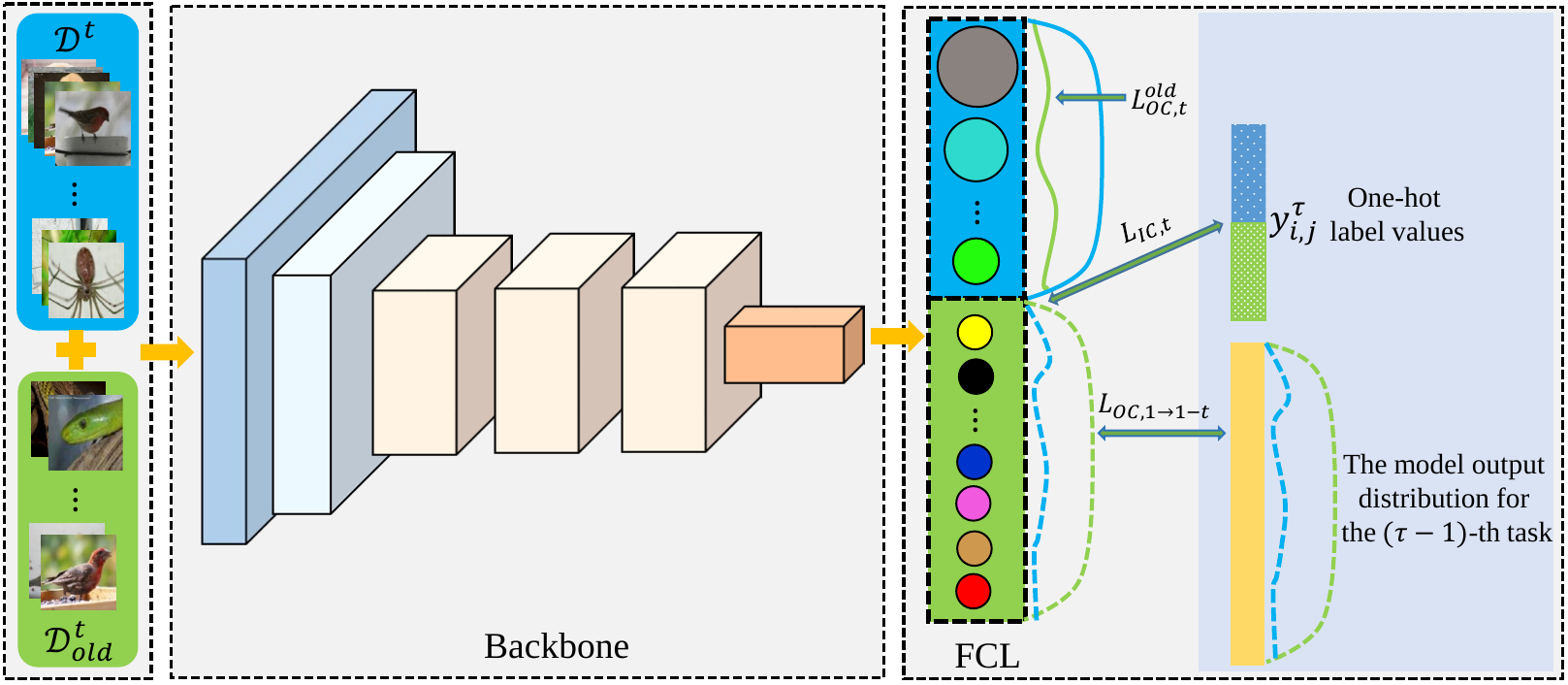}
\caption{Overall structure of the proposed method. Firstly, the absolute gradient of the output scores is computed, based on the $\hat{p}^{\tau}_{i,j,k=i}$ and the $y^{\tau}_{i,j,k=i}$, to induce a weight for each sample, where the weights are adaptively updated during the training. Then $L_{OC,1\rightarrow t-1}$ is employed to maintain the outputs for each old task. $L_{OC,1\rightarrow t-1}$ is also utilized to make the outputs of new task data on old task classification heads after updating the old task models agree with those before the update. Finally, to suppress the outputs of old task data on new task classification heads, their output scores $\hat{p}^{t}_{i,j,k}$ are directly optimized to approach zeros (The solid blue line and solid green line represents the output distribution of the new task and old task data, respectively, on the new task classification head; The dashed blue line and dashed green line represents the output distribution of the new task and old task data, respectively, on the old task classification head).}
\label{fig:overall_structure}
\end{figure*}

\section{Method}
\subsection{Notations and Problem Setup}
In CIL, data for new tasks are arriving constantly, which are represented as $\mathcal{D} =\{\mathcal{D}^1, \mathcal{D}^2, \cdots, \mathcal{D}^t, \cdots, \mathcal{D}^T \}$. The data in the $t$-th new task is $\mathcal{D}^t = \{ (x^t_{i,j}, y^t_{i,j})_{i=1,2,\cdots,m; j=1,2,\cdots,n_m} \}$, where $m$ is the number of classes, $n_m$ is the number of samples for the $m$-th category, $x$ is the input data, and $y$ is the corresponding data label.
The number of samples may vary for different categories in the new task.
When learning the $t$-th new task, we assume that there are a small amount of data stored for the old tasks, i.e.,
\begin{equation}
\label{eq:old_data_t}
\mathcal{D}^t_{old}= \left\{
(x^1_{i,j}, y^1_{i,j}), \cdots, (x^{t-1}_{i,j}, y^{t-1}_{i,j}) \right\},
\end{equation}
where $i=1,2,\cdots,m$ and $j=1,2,\cdots,n_{old}$, $n_{old} \ll n$. That is, the number of old data $\mathcal{D}^t_{old}$ in the repository is much smaller than that of $\mathcal{D}^t$. In CIL, a feature extractor $f(\cdot)$ (such as ResNet~\cite{he2016deep}) and a fully connected layer (FCL) together with a $softmax$ classifier is generally adopted, i.e.,
\begin{equation}
\label{eq:feature_extraction}
\mathbf{x}^{\tau}_{i,j}=f(x^{\tau}_{i,j}; \Theta),
\end{equation}
\begin{equation}
\label{eq:feature_classfier}
\hat{\mathbf{p}}^{\tau}_{i,j} =
softmax (\mathbf{x}^{\tau}_{i,j}; W),
\end{equation}
where $\tau = \{1, 2, \cdots, t\}$, $\Theta$ is the parameter of the feature extractor, $W$ is the parameter of the classifier, and $\hat{\mathbf{p}}^{\tau}_{i,j}$ is a vector of output scores. When incremental learning proceeds to the $t$-th task, all the data in $\mathcal{D}^t \cup \mathcal{D}^t_{old} = \left\{ (x^{\tau}_{i,j}, y^{\tau}_{i,j}), (\tau=1,2,\cdots,t) \right\}$ are utilized for training, and the following cross-entropy loss is usually adopted:
\begin{equation}
\label{eq:cross_entropy}
L_{ce,t}=-\dfrac{1}{N_{old}+n_{new}}\sum_{i,j,k,\tau=1}^{t} y^{\tau}_{i,j,k} \log(\hat{p}^{\tau}_{i,j,k}),
\end{equation}
where $N_{old}$ is total number of stored data for old tasks, $n_{new}$ is the total number of samples for the new task, and $\hat{p}^{\tau}_{i,j,k}$ is the output score at the $k$-th neuron.


\subsection{Overview}
According to the above problem setup, it can be seen that when performing incremental learning, only a limited number of samples from the old tasks will be retained. Due to the large number of samples in the new task, incremental learning suffers from the class imbalance issue between the old and new tasks. The class imbalance issue also exists within the new task, but this is ignored by existing CIL approaches~\cite{ahn2021ss,rebuffi2017icarl,yan2021dynamically}.

\begin{figure}[!t]
  \centering
  \subfloat[]
  {\label{fig:subfig1}\includegraphics[width=0.23\textwidth]{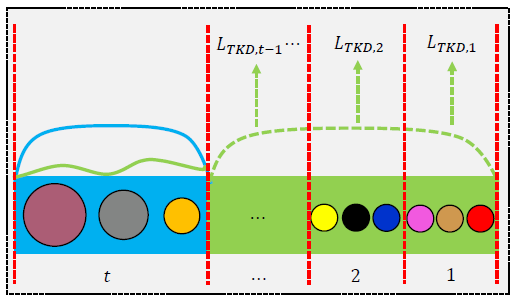}}
  \hspace{0.05cm}
  \subfloat[]
  {\label{fig:subfig2}\includegraphics[width=0.23\textwidth]{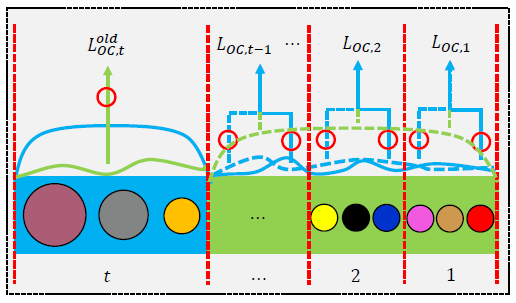}}
  \caption{A comparison of the SSIL approach (left) and the proposed output coordination (right). In SSIL, only the outputs of old task data on old classification heads are kept consistent before and after updating. We improve it by further enforcing the outputs of new task data on old classification heads to be consistent, and suppress the outputs of old task data on new classification heads.}
  \label{fig:output_coordination}
\end{figure}

Therefore, we propose the joint input and output coordination (JIOC) mechanism, as shown in Figure~\ref{fig:overall_structure}, where we assign different weights to different input data according to their the absolute value of the gradient for output scores.
In addition, in order to prevent the mutual interference of output distributions between old and new tasks, we split the softmax layer inspired by the principle of human inductive memory. This is similar to the SSIL~\cite{ahn2021ss} approach, but has several significant differences, as shown in Figure~\ref{fig:output_coordination}: 1) for each of the old tasks, we utilize KD to maintain the output distribution of each task. To make the output scores of new task data on the classification heads of old tasks consistent, we also employ KD to enforce the outputs after updating the old task models agree with the scores before the update; 2) to suppress the outputs of old task data on the classification heads of new tasks, their ground-truth target values are directly set to be zero for training.

\subsection{Input Coordination}
As we know, the class imbalance 
issue may lead to significant bias in the learned weights of the fully connected layers~\cite{li2020overcoming}.
Therefore, $\hat{p}^{\tau}_{i,j,k}$ may deviate greatly from its corresponding true value $p^{\tau}_{i,j,k}$, and hence it is necessary to balance the weight of fully connected layers between tasks and within tasks.

Due to the severe bias in the weights of the fully connected layer, we propose to utilize the 
outputs of fully connected layer's previous layer to adjust the weights.
Suppose that $\hat{\mathbf{q}}^{\tau}_{i,j}$ is the vector of the previous layer that outputs scores $\hat{\mathbf{p}}^{\tau}_{i,j}$. The derivative of $L_{ce,t}$ w.r.t. $\hat{\mathbf{q}}^{\tau}_{i,j}$ (we refer to the supplementary material for the detailed calculation) can be given by:
\begin{equation}
\label{eq:x_gradient}
\dfrac{\partial L_{ce,t}}{\partial \hat{\mathbf{q}}^{\tau}_{i,j}} = \substack{ \left[ \substack{ \hat{p}^{\tau}_{i,j,1} - y^{\tau}_{i,j,1} \\
		\cdots \\
		\hat{p}^{\tau}_{i,j,k}-y^{\tau}_{i,j,k}\\
		\cdots \\
		\hat{p}^{\tau}_{i,j, m t}-y^{\tau}_{i,j, m t}} \right]}.
\end{equation}
Then the absolute value of the gradient of the output score for the input data when $k=i$ is:
\begin{equation}
\label{eq:absolute_gradient}
\delta^{\tau}_{i,j,k=i}  = | \hat{p}^{\tau}_{i,j,k=i}-y^{\tau}_{i,j,k=i} |.
\end{equation}
When the number of data is large for a certain category, the model tends to bias to this category and thus the absolute value of the gradient in Eq.~(\ref{eq:absolute_gradient}) tends to be small in the learning process. To alleviate the bias issue, we propose to regard the absolute value as the weight for the corresponding input sample and add it into the loss during the training. That is, smaller weights will be adaptively assigned to the samples of the category that has more input data, and hence the model would focus more on the category that has fewer samples.




Based on the above analysis, we utilize the absolute values $\delta^{\tau}_{i,j}$ of the gradient to induce a weight for each input data during the training. First of all, we incorporate the absolute value of the gradient of the input data into the traditional cross-entropy loss (Eq.~(\ref{eq:cross_entropy})), i.e.,
\begin{equation}
\label{eq:input_coordination_loss}
L_{IC,t}=-\dfrac{1}{N_{old}+n_{new}}\sum_{i,j,k=i,\tau=1}^{t} y^{\tau}_{i,j,k} \delta^{\tau}_{i,j,k}\log(\hat{p}^{\tau}_{i,j,k}).
\end{equation}
Then, we can use Eq.~(\ref{eq:input_coordination_loss}) to balance the loss of each category.
In this way, the category weights of the fully connected layer can be balanced according to the absolute value $\delta_{i,j}$ of the gradient of each input data. It not only alleviates the category bias between old and new tasks in incremental learning, but also greatly reduces the within-task bias. The main procedure is summarized in Algorithm~\ref{alg1}~\footnote{In the entire algorithm pipeline, the outer loop and inner loop iterate $Total$ and $\dfrac{m*n_m+N_{old}}{batchsize}$ times, respectively. We neglect the time complexity of Eq.~(\ref{eq:absolute_gradient}), Eq.~(\ref{eq:input_coordination_loss}), as well as the parameter updates for $\Theta$ and $W$. The overall time complexity of the algorithm pipeline is $O(Total*\dfrac{m*n_m+N_{old}}{batchsize})$.}.

\begin{algorithm}[!t]
	\caption{Main procedure of input coordination.}
	\label{alg1}
	\textbf{Input}: The data of the incremental learning model $\left\{\mathcal{D}^t_{old}, \mathcal{D}^t\right\}$; the feature extractor of the current model is $f\left(\cdot, \Theta \right)$; the parameter of the current fully-connected layer is $W$; \\
        \vspace{-3ex}
        \begin{flushleft}
        \textbf{Output}: The updated parameters $\Theta$ and $W$;
        \end{flushleft}
	\begin{algorithmic}[1] 
		\FOR{$epoch = 1$; $epoch<Total$; $epoch++$ }
		\WHILE {$batchsize$ loads $\left\{\mathcal{D}^t_{old}, \mathcal{D}^t\right\}$ data}
              \STATE $\left(1\right)\delta_{i,j}^{\tau} \gets \left\{\hat{\mathbf{q}}^{\tau}_{i,j}\right\}$, by using Eq.~(\ref{eq:absolute_gradient});
            \STATE $\left(2\right)L_{IC,t} \gets L_{ce,t}$ and $\delta_{i,j}^{\tau}$, by using Eq.~(\ref{eq:input_coordination_loss});
            \STATE $\left(3\right)$ According to the loss value $L_{IC,t}$ obtained in the previous step, the parameters $\Theta$ and $W$ of the incremental learning model are updated.
		\ENDWHILE
		\STATE Return the updated $\Theta$ and $W$.
		\ENDFOR	
	\end{algorithmic}
\end{algorithm}

\subsection{Output Coordination}
According to the above analysis, it is necessary to keep the output distribution of the new task data $\mathcal{D}^t$ on the old task classification heads consistently before and after updating the old task models~\footnote{During the updating of the $\left(t-1\right)$ old tasks, there are only $m\cdot\left(t-1\right)$ classification heads. This does not contain the classification heads for the $t$-th task.}.
Also, it is necessary to suppress the output scores of the old task data $\mathcal{D}^t_{old}$ on the classification heads of the new task (In Figure~\ref{fig:class_bias}, this is to keep the blue solid line consistent with the dotted line, and make the green solid line approach to $0$).


When the model trains the $t$-th task, we suppose that the output score of the data $\mathcal{D}^t\cup \mathcal{D}^{t}_{old}$ without going through $softmax$ layer is given by $\hat{z}^{\tau}_{i,j,k}$. Before updating the old tasks models and training the $t$-th task, the output score of the data $\mathcal{D}^t \cup \mathcal{D}^{t}_{old}$ is $\tilde{ z}^{\tau}_{i,j,k}$. By considering the principle of human inductive memory, KD is used to enforce the output consistency of the new task data on the classification heads of each old task before and after updating the corresponding model, i.e.,
\begin{equation}
\label{eq:output_coordination1}
L_{OC,1\rightarrow t-1} =\sum_{\tau=1}^{t-1} \left[\sum_{i,j,k} \rho^{\epsilon}_{KL} \left(\hat{z}^{\tau}_{i,j,k}, \tilde{z}^{{\tau}}_{i,j,k}\right) \right],
\end{equation}
where $\rho^{\epsilon}_{KL}\left(\cdot\right)$ is the distillation loss, and $\epsilon$ is a temperature scaling parameter.

The output of the old task data $\mathcal{D}^{t}_{old}$ on the classification head of the new task can be adjusted according to:
\begin{equation}
\label{eq:output_coordination2}
L^{old}_{OC,t} = \frac{1}{n_{new}}\sum_{i,j,k} (\hat{p}^{t}_{i,j,k} - 0),
\end{equation}
where $i\in\left\{1, \cdots, m \left(t-1\right)\right\}$.

Although the principle of Eq.~$\left(\ref{eq:output_coordination1}\right)$ is similar to the SSIL~\cite{ahn2021ss} approach, the output coordination mechanism proposed in this paper is different from the SSIL approach, as shown in Figure~\ref{fig:output_coordination}.
Combining the output coordination loss $L_{IC,t}$ of Eq.~$\left(\ref{eq:input_coordination_loss}\right)$, the overall loss function $L_{JIOC,t}$ of the method proposed can be obtained, i.e.,
\begin{equation}
\label{eq:overall_loss}
L_{JIOC,t} = L_{IC,t} + \gamma_1 L_{OC,1\rightarrow t-1} + \gamma_2 L^{old}_{OC,t},
\end{equation}
where $\gamma_1 \geq 0 $ and $\gamma_2 \geq 0 $ are trade-off hyper-parameters.

\section{Experiment}
\subsection{Datasets and Evaluation Criteria}
\textbf{Datasets.}
In this paper, we not only validate the effectiveness of our method on unbalanced CIFAR10-LT and CIFAR100-LT datasets but also conduct corresponding validation on balanced CIFAR100~\cite{krizhevsky2009learning}, MiniImageNet~\cite{vinyals2016matching}, TinyImageNet~\cite{le2015tiny}, and Cub-200-2011~\cite{wah2011caltech} datasets. The CIFAR10 and CIFAR100 datasets both consist of $50,000$ training images and $10,000$ test images, with $10$ and $100$ categories respectively. To create unbalanced settings for the balanced datasets, we reduce the number of training samples for some classes.
To ensure that our method is applicable to various settings~\footnote{Since we use a small amount of old task data, the setup is slightly different from that of~\cite{cui2019class}.
}, we consider long-tail
imbalances~\cite{cui2019class}, and a summarization of the dataset is reported in Table~\ref{tab:imbalance_dataset}.
The MiniImageNet dataset was excerpted from the ImageNet~\cite{russakovsky2015imagenet} dataset, and it contains $100$ classes,
each with $600$ images of size $84\times84$.


\begin{table}[!t]
  \centering
    \begin{tabular}{c|c|c}
    \toprule
    \toprule
    Datasets & CIFAR10-LT & CIFAR100-LT \\
    \midrule
    Training images & 16,271 & 32,775 \\
    Classes & 10    & 100 \\
    Max \#\{images\} & 5,000 & 500 \\
    Min \#\{images\} & 206   & 200 \\
    Imbalance factor & 24    & 2.5 \\
    \bottomrule
    \bottomrule
    \end{tabular}%
    \caption{The detailed information of long-tail imbalance datasets.}
    \label{tab:imbalance_dataset}
\end{table}

Typically, the training and test split of this dataset is $80:20$.
The TinyImageNet dataset is
also a subset of the ImageNet~\cite{russakovsky2015imagenet} dataset and contains $200$ classes, with each class containing $500$ training images, $50$ validation images, and $50$ testing images. The Cub-200-2011 dataset is a bird dataset used for image classification. It covers 200 categories with a total of 11,788 images.

\textbf{Evaluation Criteria.}
Following~\cite{shi2022mimicking}, the average accuracy is used to measure the performance of the incremental learning algorithm, i.e.,
\begin{eqnarray}\label{eq20}
\bar{A} =\dfrac{1}{t} \sum_{\tau=1}^{t}A_{\tau},
\end{eqnarray}
where $A_{\tau}$ is the accuracy of the $\tau$-th task.

\begin{table*}[!t]
  \centering
  \resizebox{0.75\linewidth}{!}{
    \begin{tabular}{c|c|c|c|c|c|c}
    \toprule
    \toprule
    Dataset & \multicolumn{2}{c|}{CIFAR10-LT} & \multicolumn{4}{c}{CIFAR100-LT} \\
    \midrule
    Network & \multicolumn{4}{c|}{ResNet18} & \multicolumn{2}{c}{ResNet32} \\
    \midrule
    \midrule
    $T$ &5 &5 &10 &10 &10 &10 \\
    \midrule
    $N_{old}$ &1K &1.5K &1K &1.5K &1K &1.5K\\
    \midrule
    \midrule
    Methods & \multicolumn{6}{c}{Average accuracy} \\
    \midrule
    \midrule
    BiC~\cite{wu2019large} &65.46  &66.48 &39.91 &42.88 &60.03 &61.64\\
    \midrule
    PODNet~\cite{douillard2020podnet} &76.00 &71.23 &38.38  &41.33 &47.15 &49.21\\
    \midrule
    SSIL$^\ast$~\cite{ahn2021ss} &70.67  &73.14 &42.24  &46.14 &50.22 &55.32\\
    \midrule
    COIL~\cite{zhou2021co} &79.60 &79.64 &48.98  &50.72 &58.31 &58.31\\
    \midrule
    ICARL~\cite{rebuffi2017icarl} &70.18 &73.25 &38.47  &42.04 &50.17 &54.67\\
    \midrule
    \rowcolor[rgb]{0.753,0.753,0.753} ICARL\_JIOC  &71.37$\tiny{\pmb{+1.19}}$  &74.16$\tiny{\pmb{+0.91}}$  &46.51$\tiny{\pmb{+8.04}}$  &49.73$\tiny{\pmb{+7.69}}$ &55.24$\tiny{\pmb{+5.07}}$ &58.41$\tiny{\pmb{+3.74}}$\\
    \midrule
    DER~\cite{yan2021dynamically} &71.05  &73.27 &52.22  &54.22 &64.18 &66.13\\
    \midrule
    \rowcolor[rgb]{0.753,0.753,0.753} DER\_JIOC &71.78$\tiny{\pmb{+0.73}}$  &74.59$\tiny{\pmb{+1.32}}$ &53.20$\tiny{\pmb{+0.98}}$  &54.52$\tiny{\pmb{+0.30}}$ &66.46$\tiny{\pmb{+2.28}}$ &68.16$\tiny{\pmb{+2.03}}$ \\
    \midrule
    FOSTER~\cite{wang2022foster} &71.74  &74.39 &51.72  &49.43 &60.98 &62.82\\
    \midrule
    \rowcolor[rgb]{0.753,0.753,0.753} FOSTER\_JIOC & 73.95$\tiny{\pmb{+2.52}}$ &77.20$\tiny{\pmb{+2.72}}$ &53.28$\tiny{\pmb{+1.56}}$ &54.42$\tiny{\pmb{+4.99}}$ &62.32$\tiny{\pmb{+1.34}}$ &64.94$\tiny{\pmb{+2.12}}$ \\
    \bottomrule
    \bottomrule
    \end{tabular}%
    }
    \caption{Results on the CIFAR10-LT and CIFAR100-LT datasets ($\ast$ means our implementation).}
    \label{tb:CIFAR10-LT_and_CIFAR100-LT}
\end{table*}%

\begin{table*}[!t]
  \centering
  \resizebox{1.\linewidth}{!}{
    \begin{tabular}{c|cc|cc|c|cc|cc}
    \toprule
    \toprule
    Dataset & \multicolumn{2}{c|}{MiniImageNet} & \multicolumn{2}{c|}{TinyImageNet} & \multicolumn{1}{c|}{Cub-200-2011} & \multicolumn{4}{c}{CIFAR100}\\
    \midrule
    Network & \multicolumn{7}{c|}{ResNet18} & \multicolumn{2}{c}{ResNet32}\\
    \midrule
    \midrule
    $T$  & \multicolumn{1}{c|}{10} & 10    & \multicolumn{1}{c|}{10} & 10 & 10 & \multicolumn{1}{c|}{10} & 10  & \multicolumn{1}{c|}{10} & 10\\
    \midrule
    $N_{old}$ & \multicolumn{1}{c|}{1K} & 2K    & \multicolumn{1}{c|}{1K} & 2K & 0.5K & \multicolumn{1}{c|}{1K} & 2K & \multicolumn{1}{c|}{1K} & 2K\\
    \midrule
    \midrule
    Methods & \multicolumn{9}{c}{Average accuracy} \\
    \midrule
    \midrule
    BiC~\cite{wu2019large}&52.95 &52.10 &52.95 &54.18 &29.13 &34.43 &45.77 &58.01 &61.93\\
    \midrule
    PODNet\cite{douillard2020podnet}&55.74  &59.27 &43.30  &45.66 &31.81 &42.12  &47.25 &51.66 &53.63\\
    \midrule
    SSIL$^\ast$~\cite{ahn2021ss}&47.59  &54.52  &35.94  &42.13 &33.20 &43.63  &50.75 &51.76 &56.12\\
    \midrule
    COIL~\cite{zhou2021co}&64.97  &65.10 & 42.87 &42.84 &34.81 &51.25  &56.21 &57.12 &60.03\\
    \midrule
    ICARL~\cite{rebuffi2017icarl}&53.43  &60.90 &33.63  &39.24 &30.10 &38.29  &44.32 &52.20 &56.35\\
    \midrule
    \rowcolor[rgb]{0.753,0.753,0.753} ICARL\_JIOC&59.88$\tiny{\pmb{+6.45}}$  &65.98$\tiny{\pmb{+5.08}}$ &38.60$\tiny{\pmb{+4.97}}$  &44.74$\tiny{\pmb{+5.50}}$ &35.11$\tiny{\pmb{+5.01}}$ &47.49$\tiny{\pmb{+9.20}}$  &53.74$\tiny{\pmb{+9.42}}$ &56.66$\tiny{\pmb{+4.46}}$ &59.84$\tiny{\pmb{+3.49}}$\\
    \midrule
    DER~\cite{yan2021dynamically}&69.06  &72.36 &53.19  &56.54 &36.16 &53.16  &57.28 &67.57 &70.12\\
    \midrule
    \rowcolor[rgb]{0.753,0.753,0.753} DER\_JIOC&70.06$\tiny{\pmb{+1.00}}$   &73.08$\tiny{\pmb{+0.72}}$ &56.37$\tiny{\pmb{+3.18}}$  &57.63$\tiny{\pmb{+1.09}}$ &38.82$\tiny{\pmb{+2.26}}$ &56.29$\tiny{\pmb{+3.13}}$  &59.31$\tiny{\pmb{+2.03}}$ &70.45$\tiny{\pmb{+2.88}}$ &71.88$\tiny{\pmb{+1.76}}$\\
    \midrule
    FOSTER~\cite{wang2022foster}&67.93  &69.37 &50.36  &54.78 &27.49 &52.09  &52.09 &64.05 &65.93\\
    \midrule
    \rowcolor[rgb]{0.753,0.753,0.753} FOSTER\_JIOC&70.84$\tiny{\pmb{+2.91}}$ &73.70$\tiny{\pmb{+4.33}}$ &52.41$\tiny{\pmb{+2.05}}$ &55.91$\tiny{\pmb{+1.13}}$ &38.10$\tiny{\pmb{+10.61}}$ &55.69$\tiny{\pmb{+3.60}}$ &59.85$\tiny{\pmb{+7.76}}$ &65.12$\tiny{\pmb{+1.07}}$ &67.79$\tiny{\pmb{+1.86}}$\\
    \bottomrule
    \bottomrule
    \end{tabular}%
    }
    \caption{Results on the MiniImageNet, TinyImageNet, Cub-200, CIFAR100 datasets ($\ast$ means our impementation).}
    \label{tb:other_datasets}
\end{table*}%

\textbf{Baseline Protocol.}
The training sets of CIFAR10-LT is divided into $T=5$ tasks, and the number of categories for each task is $2$. The number of samples in the memory is fixed to be $N_{old}=\left\{1000, 2000\right\}$ during the incremental training. 
In the CIFAR100-LT, CIFAR100, MiniImageNet, TinyImageNet, and Cub-200-2011 datasets, the training tasks are divided into $T=10$. The memory sizes of CIFAR100, MiniImageNet, and TinyImageNet dataset are also fixed as $N_{old}=\left\{1000, 2000\right\}$, and the memory size is chosen from $N_{old}=\left\{1000, 1500\right\}$ In addition, for the Cub-200-2011 dataset, the memory storage size fixed as $N_{old}=\left\{500\right\}$.
The numbers of categories for each task of CIFAR10-LT, CIFAR100, MiniImageNet, TinyImageNet and Cub-200-2011 dataset are $10$, $10$, $10$, $20$ and $20$, respectively. Regarding the memory storage samples related to our fusion algorithm, we all follow the existing algorithm~\cite{rebuffi2017icarl,yan2021dynamically,wang2022foster}.

\textbf{Implementation details.}
Our method and all the compared approaches (BiC~\cite{wu2019large}, PODNet~\cite{douillard2020podnet}, COIL~\cite{zhou2021co}, SSIL~\cite{ahn2021ss}, ICARL~\cite{rebuffi2017icarl}, DER~\cite{yan2021dynamically} and FOSTER~\cite{wang2022foster}) are implemented using PyCIL~\cite{zhou2023pycil} and Pytorch~\cite{paszke2017automatic}. 

On the experimental dataset, we used ResNet18~\cite{he2016deep} and ResNet32 as feature extractors respectively. ResNet32 is just used to further demonstrate that this mechanism can have some effectiveness in other network frameworks. In terms of parameter settings, we align with the original methods on PyCIL~\cite{zhou2023pycil} to facilitate a fair comparison. Among these, the batch size is set to $128$. Additionally, the SGD optimizer is used to gradually update the weights during incremental learning model training. The learning rate is initially set to be $0.1$ and gradually decays. We run the training on two NVIDIA 3090RTX GPUs.

\subsection{Results and Analysis}
We incorporates the proposed JIOC strategy into existing class-incremental learning algorithms (ICARL~\cite{rebuffi2017icarl}, DER~\cite{yan2021dynamically}, and FOSTER~\cite{wang2022foster}). The experimental results on different datasets are shown in Table~\ref{tb:CIFAR10-LT_and_CIFAR100-LT} and Table~\ref{tb:other_datasets}.

\textbf{Results on CIFAR10-LT and CIFAR100-LT.}
From the overall performance analysis of the Table~\ref{tb:CIFAR10-LT_and_CIFAR100-LT}, it can be seen that the ICARL, DER, and FOSTER algorithms on our created imbalanced datasets have been significantly improved.
The relative improvements of ICARL\_JIOC are $1.70\%$, $1.24\%$, $20.90\%$, $18.29\%$, $10.11\%$ and $6.84\%$ compared with the original ICARL algorithm.
For the DER algorithm, our DER\_JIOC improves it by $1.03\%$, $1.80\%$, $1.88\%$, $0.55\%$, $3.55\%$ and $3.07\%$.
In regard to the FOSTER algorithm, the relative improvements are the significant $3.51\%$, $3.66\%$, $3.02\%$, $10.10\%$, $2.20\%$, and $3.37\%$ respectively.
Compared with all the counterparts, the best performance is usually achieved by the proposed FOSTER\_JIOC method.
This not only demonstrates the effectiveness of our method on imbalanced datasets,  but also further confirms its ability to alleviate catastrophic forgetting in other network frameworks.

\textbf{Results on MiniImageNet, TinyImageNet, Cub-200-2011 and CIFAR100.}
We can observe from Table~\ref{tb:other_datasets} that the mechanism proposed in this paper also has significant improvements on the MiniImageNet, TinyImageNet, Cub-200-2011 and CIFAR100 datasets.
For example, 
our FOSTER\_JIOC outperforms the original FOSTER algorithm by $4.28\%$, $6.24\%$, $4.07\%$, $2.06\%$, $38.60\%$, $6.91\%$, $14.90\%$, $1.67\%$ and $2.82\%$, respectively.
The best performance is also achieved by the proposed DER\_JIOC and ICARL\_JIOC, which are comparable. This further demonstrates that the proposed method not only alleviates catastrophic forgetting in class-imbalanced datasets but also has a forgetting-mitigation effect on normal data.

\subsection{Ablation Studies}
In this section, we first separately investigate the effectiveness of input and output coordination strategies, and then analyze that the proposed output coordination strategy exhibits a more pronounced effect in mitigating forgetting compared to the SSIL method.

\begin{table*}[!t]
  \centering
  \resizebox{0.9\linewidth}{!}{
    \begin{tabular}{c|c|c|c|c|c|c|c|c|c|c|c|c}
    \toprule
    \toprule
    \multicolumn{2}{c}{\multirow{2}[4]{*}{ICARL}} & \multicolumn{11}{c}{Task} \\
\cmidrule{3-13}    \multicolumn{2}{c}{} & 1     & 2     & 3     & 4     & 5     & 6     & 7     & 8     & 9     & \multicolumn{1}{c}{10} & Avg \\
    \midrule
    \midrule
    \multirow{4}[2]{*}{$All_{Tasks}$} & $L_{ce}+L_{KD}$ & 80.70  & 58.95 & 50.97 & 39.60  & 35.06 & 30.45 & 29.07 & 22.15 & 19.57 & 18.21 & 38.47 \\
          & $\pmb{L_{IC}}+L_{KD}$    & 80.70  & 63.50  & 55.80  & 45.18 & 40.84 & 36.88 & 33.46 & 27.21 & 25.24 & 22.79 & 43.16 \\
          & $L_{ce}+\pmb{L_{OC}}$    & 80.70  & 63.75 & 57.50  & 46.90  & 43.58 & 38.07 & 36.56 & 30.71 & 28.56 & 24.98 & 45.13 \\
          & $\pmb{L_{JIOC}}$  & 80.70  & 64.90  & 58.37 & 47.50  & 46.08 & 40.33 & 36.79 & 32.66 & 30.46 & 27.26 & \textbf{46.51} \\
    \midrule
    \multirow{4}[2]{*}{$New_{Task}$} & $L_{ce}+L_{KD}$ & 80.70 & 61.50  & 76.1  & 70.20  & 81.9  & 70.01 & 74.7  & 73.00    & 77.10  & 71.20  & 73.64 \\
          & $\pmb{L_{IC}}+L_{KD}$   & 80.70  & 66.40  & 81.00    & 73.10  & 84.60  & 72.50  & 77.7  & 74.90  & 79.70  & 71.60  & \textbf{76.22} \\
          & $L_{ce}+\pmb{L_{OC}}$    & 80.70  & 64.40  & 80.30  & 71.70  & 83.20  & 69.70  & 76.50  & 72.40  & 78.80  & 70.60  & 74.83 \\
          & $\pmb{L_{JIOC}}$  & 80.70 & 66.40  & 80.40  & 70.90  & 84.20  & 71.50  & 80.30  & 73.10  & 78.40  & 72.40  & 75.83 \\
    \midrule
    \multirow{4}[2]{*}{$Old_{Tasks}$} & $L_{ce}+L_{KD}$ & \rule[2pt]{0.9em}{0.05em}     & 56.40  & 38.40  & 29.40  & 23.35 & 22.52 & 19.15 & 14.89 & 12.38 & 12.32 & 25.42 \\
          & $\pmb{L_{IC}}+L_{KD}$    & \rule[2pt]{0.9em}{0.05em}     & 60.60  & 45.35 & 37.43 & 32.52 & 30.28 & 25.65 & 21.89 & 19.30  & 18.00    & 32.34 \\
          & $L_{ce}+\pmb{L_{OC}}$   & \rule[2pt]{0.9em}{0.05em}     & 63.10  & 46.10  & 38.63 & 33.67 & 31.74 & 29.90  & 24.76 & 22.28 & 19.91 & 34.45 \\
          & $\pmb{L_{JIOC}}$  & \rule[2pt]{0.9em}{0.05em}     & 63.40  & 47.35 & 39.70  & 36.55 & 34.1  & 29.53 & 26.89 & 24.46 & 22.24 & \textbf{36.02} \\
    \bottomrule
    \bottomrule
    \end{tabular}}%
    \caption{The results obtained by running with $N_{old}=1000$, using ResNet18 as the feature extractor on the CIFAR100-LT dataset ($L_{ce}+L_{KD}$ is the loss function used by the ICARL algorithm).}
    \label{tb:CIFAR100-LT-res18}
\end{table*}%

\begin{figure*}[!t]
  \centering
  \captionsetup[subfloat]{font=footnotesize,labelfont=rm,textfont=rm}
  \subfloat[The results of using ResNet18 as the feature extractor on CIFAR100-LT.]
  {\label{fig:stdy1}\includegraphics[scale=0.27]{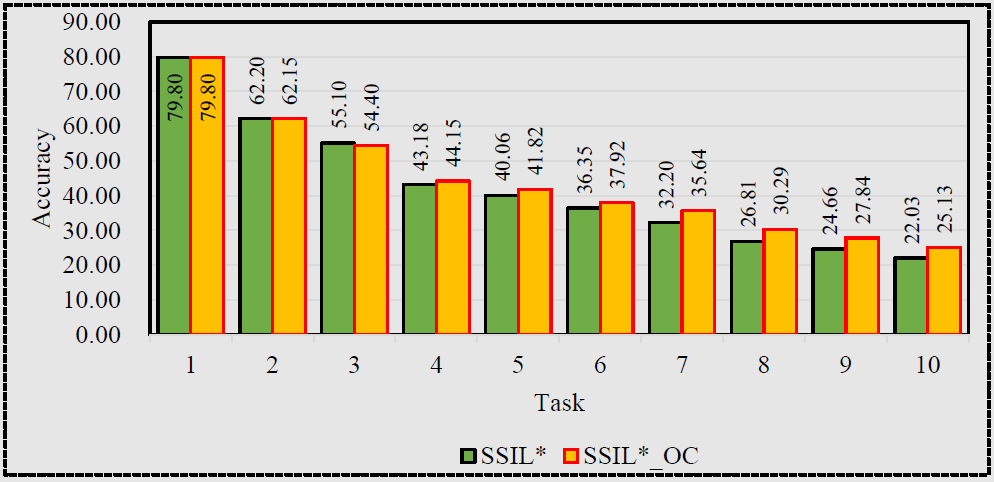}}
  \hspace{0.5mm}
  \subfloat[The results of using ResNet18 as the feature extractor on CIFAR100.]
  {\label{fig:stdy1}\includegraphics[scale=0.27]{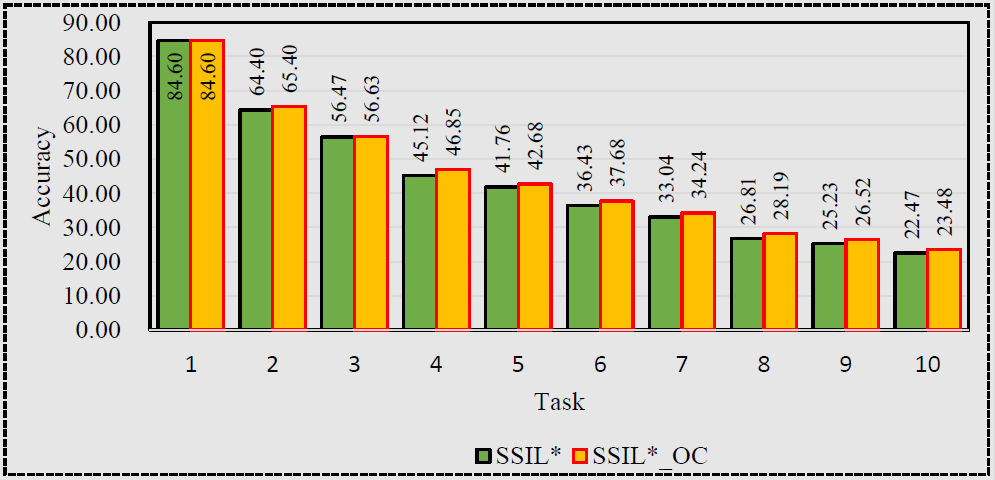}}
\hspace{0.5mm}
  \subfloat[The results of using ResNet32 as the feature extractor on CIFAR100.]
  {\label{fig:stdy1}\includegraphics[scale=0.27]{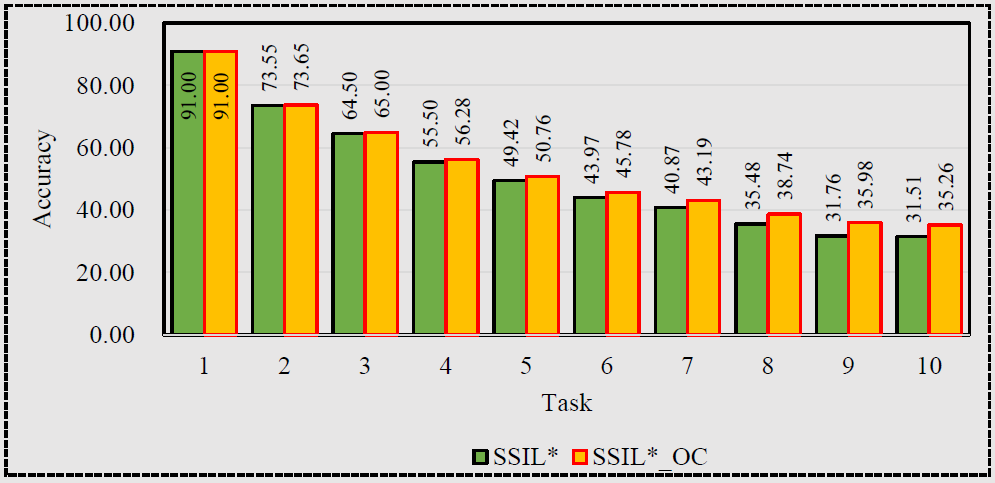}}
  \caption{A Compare the experimental results of SSIL with those of SSIL\_OC (SSIL\_OC is formed by integrating the proposed OC strategy into the SSIL algorithm. $\ast$ means our impementation).}
  \label{fig:ssil_and_ssil_oc}
\end{figure*}

\textbf{Study on the effectiveness of Input and Output Coordination.} 
To validate the effectiveness of input and output coordination strategies, we exclusively employed ResNet18 as the feature extractor on the CIFAR100-LT dataset, , as shown in Tables~\ref{tb:CIFAR100-LT-res18}. Furthermore, on the CIFAR100 dataset, we conducted experiments separately using ResNet18 and ResNet32 as feature extractors, as shown in Table 5 and Table 6 (we refer to the supplementary material for Table 5 and Table 6). 

$\left(1\right)$ The results are reported in Table~\ref{tb:CIFAR100-LT-res18}, where we can see that the average accuracy of the ICARL algorithm with the proposed input coordination is $32.34$ on the old tasks, $76.22$ on the new task, and $43.16$ overall.
Compared with the original ICARL approach, the improvements are $27.22\%$, $3.50\%$, and $12.19\%$, respectively. The results from the old task, new task, and overall task in Table 5 and Table 6 also illustrate the competitiveness of the input coordination strategy. This demonstrates that the input coordination strategy can alleviate class imbalance in incremental learning.
Besides, the ICARL algorithm only uses KD to maintain the output distribution on the old task classification heads. It does not take into account the human inductive memory mechanism for coordinating output distribution across different tasks. In Table~\ref{tb:CIFAR100-LT-res18}, the ICARL algorithm achieved average performances of $34.45$, $74.83$, and $45.13$ on the old tasks, new tasks, and overall, respectively. Our proposed output coordination mechanisms improved these performances by $35.52\%$, $1.62\%$, and $17.31\%$, respectively.
It can be concluded that the input and output coordination strategy proposed in this paper yields significant improvements, whether applied to the CIAFR100 dataset, the CIAFR100-LT dataset, or different network architectures with varying depths.

$\left(2\right)$ In the CIART100-LT dataset, the input coordination strategy demonstrates notable enhancements in the outcomes for old tasks $\left(27.22\%\right)$, new tasks $\left(3.50\%\right)$, and overall task $\left(12.19\%\right)$ performance when compared to the original ICARL algorithm, as shown in Table~\ref{tb:CIFAR100-LT-res18}. Similarly, In the CIART100 dataset, the input coordination strategy further improves the original ICARL algorithm by 42.05\%, 3.61\%, and 19.51\% on the old tasks, new tasks, and overall tasks, as shown in Table 5. According to the description and corresponding improvement effects of CIART100-LT and CIART100, it can be seen that the input coordination strategy has a good regulating effect on the imbalance of old and new categories. 



\textbf{Experiment Comparison between Output Coordination Strategy and the SSIL.} To quantitatively analyze the differences between the output coordination strategy and the SSIL, we conducted corresponding experimental results based on different feature extractors and datasets, including class-imbalanced and balanced datasets, as shown in Figure~\ref{fig:ssil_and_ssil_oc}. From the results in Figure~\ref{fig:ssil_and_ssil_oc}, it is evident that using the output distribution coordination strategy leads to significant improvements in each stage task on class-imbalanced datasets and in deep feature networks (ResNet32). This also indicates that the output distribution coordination strategy enables new task data to maintain consistent output distributions on the old task classification head and suppresses old task data on the new task classification head during the incremental learning process. This avoidance of interference between new and old tasks is achieved. However, SSIL does not avoid interference from old task outputs, which results in its performance being inferior to the output distribution coordination strategy.

\section{Conclusion}
Although the existing approaches address the class bias issue in class-incremental learning (CIL) to a certain extent by scaling and dividing the softmax layer, they all ignore the bias within the task. In addition, the mutual interference between old and new tasks has not been well resolved.
Therefore, we propose a joint input and output coordination (JIOC) mechanism to enable incremental learning models to simultaneously reduce the interference between predictions for these tasks and alleviate the class imbalance issue between and within tasks. From the extensive experiments on multiple popular datasets, we observe significant improvements when incorporating the proposed mechanism into the existing CIL approaches that utilize memory storage.
In the future, we intend to design more sophisticated strategies to reweight the inputs, and develop a general end-to-end framework for CIL.

\subsubsection{Acknowledgements}
This work is supported in part by the National Natural Science Foundation of China (Grant No. U23A20318), the Special Fund of Hubei Luojia Laboratory under Grant 220100014, the Fundamental Research Funds for the Central Universities (No. 2042024kf0039), the National Research Foundation Singapore and DSO National Laboratories under the AI Singapore Programme (AISG Award No: AISG2-GC-2023-006), and the CCF-Zhipu AI Large Model Fund OF 202224.

\bibliographystyle{named}
\bibliography{ijcai24}

\begin{thebibliography}{}

\bibitem[\protect\citeauthoryear{Ahn \bgroup \em et al.\egroup }{2021}]{ahn2021ss}
Hongjoon Ahn, Jihwan Kwak, Subin Lim, Hyeonsu Bang, Hyojun Kim, and Taesup Moon.
\newblock Ss-il: Separated softmax for incremental learning.
\newblock In {\em Proceedings of the IEEE/CVF International conference on computer vision}, pages 844--853, 2021.

\bibitem[\protect\citeauthoryear{Aljundi \bgroup \em et al.\egroup }{2019}]{aljundi2019task}
Rahaf Aljundi, Klaas Kelchtermans, and Tinne Tuytelaars.
\newblock Task-free continual learning.
\newblock In {\em Proceedings of the IEEE/CVF Conference on Computer Vision and Pattern Recognition}, pages 11254--11263, 2019.

\bibitem[\protect\citeauthoryear{Cheraghian \bgroup \em et al.\egroup }{2021}]{cheraghian2021synthesized}
Ali Cheraghian, Shafin Rahman, Sameera Ramasinghe, Pengfei Fang, Christian Simon, Lars Petersson, and Mehrtash Harandi.
\newblock Synthesized feature based few-shot class-incremental learning on a mixture of subspaces.
\newblock In {\em Proceedings of the IEEE/CVF International Conference on Computer Vision}, pages 8661--8670, 2021.

\bibitem[\protect\citeauthoryear{Cui \bgroup \em et al.\egroup }{2019}]{cui2019class}
Yin Cui, Menglin Jia, Tsung-Yi Lin, Yang Song, and Serge Belongie.
\newblock Class-balanced loss based on effective number of samples.
\newblock In {\em Proceedings of the IEEE/CVF conference on computer vision and pattern recognition}, pages 9268--9277, 2019.

\bibitem[\protect\citeauthoryear{De~Lange \bgroup \em et al.\egroup }{2021}]{de2021continual}
Matthias De~Lange, Rahaf Aljundi, Marc Masana, Sarah Parisot, Xu~Jia, Ale{\v{s}} Leonardis, Gregory Slabaugh, and Tinne Tuytelaars.
\newblock A continual learning survey: Defying forgetting in classification tasks.
\newblock {\em IEEE transactions on pattern analysis and machine intelligence}, 44(7):3366--3385, 2021.

\bibitem[\protect\citeauthoryear{Douillard \bgroup \em et al.\egroup }{2020}]{douillard2020podnet}
Arthur Douillard, Matthieu Cord, Charles Ollion, Thomas Robert, and Eduardo Valle.
\newblock Podnet: Pooled outputs distillation for small-tasks incremental learning.
\newblock In {\em Computer Vision--ECCV 2020: 16th European Conference, Glasgow, UK, August 23--28, 2020, Proceedings, Part XX 16}, pages 86--102. Springer, 2020.

\bibitem[\protect\citeauthoryear{Fisher and Sloutsky}{2005}]{fisher2005induction}
Anna~V Fisher and Vladimir~M Sloutsky.
\newblock When induction meets memory: Evidence for gradual transition from similarity-based to category-based induction.
\newblock {\em Child development}, 76(3):583--597, 2005.

\bibitem[\protect\citeauthoryear{French and Chater}{2002}]{french2002using}
Robert~M French and Nick Chater.
\newblock Using noise to compute error surfaces in connectionist networks: A novel means of reducing catastrophic forgetting.
\newblock {\em Neural computation}, 14(7):1755--1769, 2002.

\bibitem[\protect\citeauthoryear{Garg \bgroup \em et al.\egroup }{2022}]{garg2022multi}
Prachi Garg, Rohit Saluja, Vineeth~N Balasubramanian, Chetan Arora, Anbumani Subramanian, and CV~Jawahar.
\newblock Multi-domain incremental learning for semantic segmentation.
\newblock In {\em Proceedings of the IEEE/CVF Winter Conference on Applications of Computer Vision}, pages 761--771, 2022.

\bibitem[\protect\citeauthoryear{Geng \bgroup \em et al.\egroup }{2020}]{geng2020dynamic}
Ruiying Geng, Binhua Li, Yongbin Li, Jian Sun, and Xiaodan Zhu.
\newblock Dynamic memory induction networks for few-shot text classification.
\newblock {\em arXiv preprint arXiv:2005.05727}, 2020.

\bibitem[\protect\citeauthoryear{Hayes \bgroup \em et al.\egroup }{2013}]{Hayes2013}
Brett~K Hayes, Kristina Fritz, and Evan Heit.
\newblock The relationship between memory and inductive reasoning: Does it develop ?
\newblock In {\em Developmental Psychology}, volume~44, pages 848--860, 2013.

\bibitem[\protect\citeauthoryear{He \bgroup \em et al.\egroup }{2016}]{he2016deep}
Kaiming He, Xiangyu Zhang, Shaoqing Ren, and Jian Sun.
\newblock Deep residual learning for image recognition.
\newblock In {\em Proceedings of the IEEE conference on computer vision and pattern recognition}, pages 770--778, 2016.

\bibitem[\protect\citeauthoryear{Krizhevsky \bgroup \em et al.\egroup }{2009}]{krizhevsky2009learning}
Alex Krizhevsky, Geoffrey Hinton, et~al.
\newblock Learning multiple layers of features from tiny images.
\newblock 2009.

\bibitem[\protect\citeauthoryear{Le and Yang}{2015}]{le2015tiny}
Ya~Le and Xuan Yang.
\newblock Tiny imagenet visual recognition challenge.
\newblock {\em CS 231N}, 7(7):3, 2015.

\bibitem[\protect\citeauthoryear{Li and Hoiem}{2017}]{li2017learning}
Zhizhong Li and Derek Hoiem.
\newblock Learning without forgetting.
\newblock {\em IEEE transactions on pattern analysis and machine intelligence}, 40(12):2935--2947, 2017.

\bibitem[\protect\citeauthoryear{Li \bgroup \em et al.\egroup }{2020}]{li2020overcoming}
Yu~Li, Tao Wang, Bingyi Kang, Sheng Tang, Chunfeng Wang, Jintao Li, and Jiashi Feng.
\newblock Overcoming classifier imbalance for long-tail object detection with balanced group softmax.
\newblock In {\em Proceedings of the IEEE/CVF conference on computer vision and pattern recognition}, pages 10991--11000, 2020.

\bibitem[\protect\citeauthoryear{Liu \bgroup \em et al.\egroup }{2021}]{liu2021adaptive}
Yaoyao Liu, Bernt Schiele, and Qianru Sun.
\newblock Adaptive aggregation networks for class-incremental learning.
\newblock In {\em Proceedings of the IEEE/CVF conference on Computer Vision and Pattern Recognition}, pages 2544--2553, 2021.

\bibitem[\protect\citeauthoryear{Mallya and Lazebnik}{2018}]{mallya2018packnet}
Arun Mallya and Svetlana Lazebnik.
\newblock Packnet: Adding multiple tasks to a single network by iterative pruning.
\newblock In {\em Proceedings of the IEEE conference on Computer Vision and Pattern Recognition}, pages 7765--7773, 2018.

\bibitem[\protect\citeauthoryear{Mallya \bgroup \em et al.\egroup }{2018}]{mallya2018piggyback}
Arun Mallya, Dillon Davis, and Svetlana Lazebnik.
\newblock Piggyback: Adapting a single network to multiple tasks by learning to mask weights.
\newblock In {\em Proceedings of the European Conference on Computer Vision (ECCV)}, pages 67--82, 2018.

\bibitem[\protect\citeauthoryear{Menon \bgroup \em et al.\egroup }{2021}]{menon2021statistical}
Aditya~K Menon, Ankit~Singh Rawat, Sashank Reddi, Seungyeon Kim, and Sanjiv Kumar.
\newblock A statistical perspective on distillation.
\newblock In {\em International Conference on Machine Learning}, pages 7632--7642. PMLR, 2021.

\bibitem[\protect\citeauthoryear{Mirza \bgroup \em et al.\egroup }{2022}]{mirza2022efficient}
M~Jehanzeb Mirza, Marc Masana, Horst Possegger, and Horst Bischof.
\newblock An efficient domain-incremental learning approach to drive in all weather conditions.
\newblock In {\em Proceedings of the IEEE/CVF Conference on Computer Vision and Pattern Recognition}, pages 3001--3011, 2022.

\bibitem[\protect\citeauthoryear{Paszke \bgroup \em et al.\egroup }{2017}]{paszke2017automatic}
Adam Paszke, Sam Gross, Soumith Chintala, Gregory Chanan, Edward Yang, Zachary DeVito, Zeming Lin, Alban Desmaison, Luca Antiga, and Adam Lerer.
\newblock Automatic differentiation in pytorch.
\newblock 2017.

\bibitem[\protect\citeauthoryear{Rebuffi \bgroup \em et al.\egroup }{2017}]{rebuffi2017icarl}
Sylvestre-Alvise Rebuffi, Alexander Kolesnikov, Georg Sperl, and Christoph~H Lampert.
\newblock icarl: Incremental classifier and representation learning.
\newblock In {\em Proceedings of the IEEE conference on Computer Vision and Pattern Recognition}, pages 2001--2010, 2017.

\bibitem[\protect\citeauthoryear{Redondo and Morris}{2011}]{redondo2011making}
Roger~L Redondo and Richard~GM Morris.
\newblock Making memories last: the synaptic tagging and capture hypothesis.
\newblock {\em Nature Reviews Neuroscience}, 12(1):17--30, 2011.

\bibitem[\protect\citeauthoryear{Russakovsky \bgroup \em et al.\egroup }{2015}]{russakovsky2015imagenet}
Olga Russakovsky, Jia Deng, Hao Su, Jonathan Krause, Sanjeev Satheesh, Sean Ma, Zhiheng Huang, Andrej Karpathy, Aditya Khosla, Michael Bernstein, et~al.
\newblock Imagenet large scale visual recognition challenge.
\newblock {\em International journal of computer vision}, 115:211--252, 2015.

\bibitem[\protect\citeauthoryear{Santoso and Finn}{2022}]{santoso2022data}
Fendy Santoso and Anthony Finn.
\newblock A data-driven cyber--physical system using deep-learning convolutional neural networks: Study on false-data injection attacks in an unmanned ground vehicle under fault-tolerant conditions.
\newblock {\em IEEE Transactions on Systems, Man, and Cybernetics: Systems}, 53(1):346--356, 2022.

\bibitem[\protect\citeauthoryear{Serra \bgroup \em et al.\egroup }{2018}]{serra2018overcoming}
Joan Serra, Didac Suris, Marius Miron, and Alexandros Karatzoglou.
\newblock Overcoming catastrophic forgetting with hard attention to the task.
\newblock In {\em International Conference on Machine Learning}, pages 4548--4557. PMLR, 2018.

\bibitem[\protect\citeauthoryear{Shi \bgroup \em et al.\egroup }{2022}]{shi2022mimicking}
Yujun Shi, Kuangqi Zhou, Jian Liang, Zihang Jiang, Jiashi Feng, Philip~HS Torr, Song Bai, and Vincent~YF Tan.
\newblock Mimicking the oracle: an initial phase decorrelation approach for class incremental learning.
\newblock In {\em Proceedings of the IEEE/CVF Conference on Computer Vision and Pattern Recognition}, pages 16722--16731, 2022.

\bibitem[\protect\citeauthoryear{Tao \bgroup \em et al.\egroup }{2020}]{tao2020few}
Xiaoyu Tao, Xiaopeng Hong, Xinyuan Chang, Songlin Dong, Xing Wei, and Yihong Gong.
\newblock Few-shot class-incremental learning.
\newblock In {\em Proceedings of the IEEE/CVF Conference on Computer Vision and Pattern Recognition}, pages 12183--12192, 2020.

\bibitem[\protect\citeauthoryear{Tschandl \bgroup \em et al.\egroup }{2020}]{tschandl2020human}
Philipp Tschandl, Christoph Rinner, Zoe Apalla, Giuseppe Argenziano, Noel Codella, Allan Halpern, Monika Janda, Aimilios Lallas, Caterina Longo, Josep Malvehy, et~al.
\newblock Human--computer collaboration for skin cancer recognition.
\newblock {\em Nature Medicine}, 26(8):1229--1234, 2020.

\bibitem[\protect\citeauthoryear{Vinyals \bgroup \em et al.\egroup }{2016}]{vinyals2016matching}
Oriol Vinyals, Charles Blundell, Timothy Lillicrap, Daan Wierstra, et~al.
\newblock Matching networks for one shot learning.
\newblock {\em Advances in neural information processing systems}, 29, 2016.

\bibitem[\protect\citeauthoryear{Wah \bgroup \em et al.\egroup }{2011}]{wah2011caltech}
Catherine Wah, Steve Branson, Peter Welinder, Pietro Perona, and Serge Belongie.
\newblock The caltech-ucsd birds-200-2011 dataset.
\newblock 2011.

\bibitem[\protect\citeauthoryear{Wang \bgroup \em et al.\egroup }{2022}]{wang2022foster}
Fu-Yun Wang, Da-Wei Zhou, Han-Jia Ye, and De-Chuan Zhan.
\newblock Foster: Feature boosting and compression for class-incremental learning.
\newblock In {\em Computer Vision--ECCV 2022: 17th European Conference, Tel Aviv, Israel, October 23--27, 2022, Proceedings, Part XXV}, pages 398--414. Springer, 2022.

\bibitem[\protect\citeauthoryear{Williams}{1999}]{williams1999memory}
John~N Williams.
\newblock Memory, attention, and inductive learning.
\newblock {\em Studies in Second Language Acquisition}, 21(1):1--48, 1999.

\bibitem[\protect\citeauthoryear{Wu \bgroup \em et al.\egroup }{2019}]{wu2019large}
Yue Wu, Yinpeng Chen, Lijuan Wang, Yuancheng Ye, Zicheng Liu, Yandong Guo, and Yun Fu.
\newblock Large scale incremental learning.
\newblock In {\em Proceedings of the IEEE/CVF Conference on Computer Vision and Pattern Recognition}, pages 374--382, 2019.

\bibitem[\protect\citeauthoryear{Yan \bgroup \em et al.\egroup }{2021}]{yan2021dynamically}
Shipeng Yan, Jiangwei Xie, and Xuming He.
\newblock Der: Dynamically expandable representation for class incremental learning.
\newblock In {\em Proceedings of the IEEE/CVF Conference on Computer Vision and Pattern Recognition}, pages 3014--3023, 2021.

\bibitem[\protect\citeauthoryear{Zhang \bgroup \em et al.\egroup }{2020}]{zhang2020class}
Junting Zhang, Jie Zhang, Shalini Ghosh, Dawei Li, Serafettin Tasci, Larry Heck, Heming Zhang, and C-C~Jay Kuo.
\newblock Class-incremental learning via deep model consolidation.
\newblock In {\em Proceedings of the IEEE/CVF Winter Conference on Applications of Computer Vision}, pages 1131--1140, 2020.

\bibitem[\protect\citeauthoryear{Zhou \bgroup \em et al.\egroup }{2021}]{zhou2021co}
Da-Wei Zhou, Han-Jia Ye, and De-Chuan Zhan.
\newblock Co-transport for class-incremental learning.
\newblock In {\em Proceedings of the 29th ACM International Conference on Multimedia}, pages 1645--1654, 2021.

\bibitem[\protect\citeauthoryear{Zhou \bgroup \em et al.\egroup }{2023}]{zhou2023pycil}
Da-Wei Zhou, Fu-Yun Wang, Han-Jia Ye, and De-Chuan Zhan.
\newblock Pycil: a python toolbox for class-incremental learning.
\newblock {\em SCIENCE CHINA Information Sciences}, 66(9):197101--, 2023.

\end{thebibliography}
\clearpage
\appendix
\section*{\centerline{Appendix}}
\section{Absolute Value of Gradient}
According to the above problem setup, the data $\left\{ (x^{t^{\prime}}_{i,j}, y^{t^{\prime}}_{i,j}) \right\}$ in the $t$-th task is simplified to $\left\{ (x_{u}, y_{u}), (u=1,2,\cdots,m \cdot t) \right\}$, where $m \cdot t$ is the number of classes. Besides, the corresponding output score $\hat{\mathbf{p}}^{t^{\prime}}_{i,j}$ is simplified to $\hat{\mathbf{p}}_{u}$.
The previous layer's output score for softmax is $\hat{\mathbf{q}}_{u}$, i.e.,
\begin{equation}\label{eq:softmax}
\hat{\mathbf{p}}_{u}=\dfrac{e^{\hat{\mathbf{q}}_{u}}}{\sum_{r=1}^{m \cdot t}e^{\hat{\mathbf{q}}_{r}}}
\end{equation}

 If the $k$-th neuron is the correct output label, $y_{k}=1$ in $\left[y_{1},y_{,2},\cdots,y_{m \cdot t}\right]$ and others are 0. The derivative of $L_{ce,t}$ w.r.t.  $\hat{\mathbf{q}}_{u}$ can be given by:
\begin{equation}\label{eq:derivation}
\begin{split}
\dfrac{\vartheta L_{ce,t}}{\vartheta \hat{\mathbf{q}}_{u}}
&=\dfrac{\vartheta L_{ce,t}}{\vartheta \hat{\mathbf{p}}_{u}} \cdot \dfrac{\vartheta \hat{\mathbf{p}}_{u}}{\vartheta \hat{\mathbf{q}}_{u}}\\
&=\dfrac{\vartheta \left(-\sum_{u=1}^{m \cdot t} y_{u}log{\hat{\mathbf{p}}_{u}}\right)}{\vartheta \hat{\mathbf{p}}_{u}} \cdot \dfrac{\vartheta\hat{\mathbf{p}}_{u}}{\vartheta\hat{\mathbf{q}}_{u}}\\
\end{split}
\end{equation}

Since $y_k=1$, $\dfrac{\vartheta L_{ce,t}}{\vartheta \hat{p}_{u}} \neq 0$, and the others are 0. The above Eq. \ref{eq:derivation} can be further simplified as:
\begin{equation}\label{eq:derivation_simplified}
\begin{split}
\dfrac{\vartheta L_{ce,t}}{\vartheta \hat{\mathbf{q}}_{u}}
&=\dfrac{y_{u}}{\hat{p}_{k}} \cdot \dfrac{\vartheta\hat{p}_{k}}{\vartheta\hat{\mathbf{q}}_{u}} \\
\end{split}
\end{equation}

The solution of $\dfrac{\vartheta L_{ce,t}}{\vartheta \hat{\mathbf{q}}_{u}}$ in Eq. \ref{eq:derivation_simplified} needs to be divided into two cases ($u=k$ and $u \neq k$).

$u=k$,
\begin{equation}\label{eq:u_equal_k}
\begin{split}
\dfrac{\vartheta\hat{p}_{k}}{\vartheta\hat{\mathbf{q}}_{u}}&=\dfrac{\vartheta\hat{p}_{k}}{\vartheta\hat{q}_{k}}\\
&=\dfrac{e^{\hat{q}_k}}{\sum_{r=1}^{m \cdot t}e^{\hat{q}_r}}-\left(\dfrac{e^{\hat{q}_k}}{\sum_{r=1}^{m \cdot t}e^{\hat{q}_r}}\right)^2 \\
&=\hat{p}_k\left(1-\hat{p}_k\right)
\end{split}
\end{equation}

$u\neq k$,
\begin{equation}\label{eq:u_notequal_k}
\begin{split}
\dfrac{\vartheta\hat{p}_{k}}{\vartheta\hat{\mathbf{q}}_{u}}=-\hat{p}_k\hat{\mathbf{p}}_u
\end{split}
\end{equation}

From the above Eq. \ref{eq:u_equal_k} and \ref{eq:u_notequal_k}, The derivative of $L_{ce,t}$ w.r.t.  $\hat{\mathbf{q}}_{u}$ can be given by:
\begin{equation}\label{eq:last_derivation}
\begin{split}
\dfrac{\vartheta L_{ce,t}}{\vartheta \hat{\mathbf{q}}_{u}}
&=\left[
\begin{matrix}
\hat{p}_1\\
\cdots \\
\hat{p}_k-1\\
\cdots \\
\hat{p}_{m \cdot t}
\end{matrix}
\right]
\end{split}
\end{equation}

Since $y_k=1$ and others are 0, the above Eq. \ref{eq:last_derivation} can be further rewritten, i.e.,
\begin{equation}\label{eq:re_last_derivation}
\begin{split}
\dfrac{\vartheta L_{ce,t}}{\vartheta \hat{\mathbf{q}}_{u}}&=\left[
\begin{matrix}
\hat{p}_1-y_1 \\
\cdots \\
\hat{p}_k-y_k \\
\cdot  \\
\hat{p}_{m \cdot t}-y_{m \cdot t}
\end{matrix}
\right]
\end{split}
\end{equation}

\section{Ablation Studies Table}
On the CIFAR100 dataset, we conducted experiments separately using ResNet18 and ResNet32 as feature extractors, as shown in Table~\ref{tb:CIFAR100_res18} and Table~\ref{tb:CIFAR100_res32}.
\begin{table*}[!t]
  \centering
  \resizebox{1.\linewidth}{!}{
    \begin{tabular}{c|c|c|c|c|c|c|c|c|c|c|c|c}
    \toprule
    \toprule
    \multicolumn{2}{c}{\multirow{2}[4]{*}{ICARL}} & \multicolumn{11}{c}{Task} \\
\cmidrule{3-13}    \multicolumn{2}{c}{} & 1     & 2     & 3     & 4     & 5     & 6     & 7     & 8     & 9     & \multicolumn{1}{c}{10} & Avg \\
    \midrule
    \midrule
    \multirow{4}[2]{*}{$All_{Tasks}$} & $L_{ce}+L_{KD}$ &87.10	&60.75	&49.13	&39.22	&33.54	&29.03	&26.40	&21.31	&19.54	&16.83	&38.29\\
          & $\pmb{L_{IC}}+L_{KD}$ &87.10	&68.65	&59.07	&48.12	&42.36	&37.00	&35.37	&29.65	&26.94	&23.34	&45.76\\
          & $L_{ce}+\pmb{L_{OC}}$  &87.10	&69.15	&59.57	&48.42	&45.10	&40.98	&39.23	&32.48	&30.26	&27.55	&47.98 \\
          & $\pmb{L_{JIOC}}$  &87.10	&68.80	&58.93	&49.55	&45.96	&40.68	&38.73	&33.39	&29.86	&27.88	&\textbf{48.09} \\
    \midrule
    \multirow{4}[2]{*}{$New_{Task}$} & $L_{ce}+L_{KD}$ &87.10	&66.80	&80.70	&74.70	&85.10	&75.40	&79.10	&76.40	&79.50	&73.40	&77.82\\
          & $\pmb{L_{IC}}+L_{KD}$   &87.10	&71.80	&85.20	&77.30	&86.20	&77.90	&82.70	&79.90	&82.40	&75.80	&\textbf{80.63}\\
          & $L_{ce}+\pmb{L_{OC}}$    &87.10	&71.90	&84.50	&76.00	&85.50	&77.70	&82.50	&78.40	&82.20	&76.40	&80.22\\
          & $\pmb{L_{JIOC}}$  &87.10	&71.70	&84.10	&77.40	&85.40	&76.30	&81.70	&79.30	&82.20	&77.20	&80.24\\
    \midrule
    \multirow{4}[2]{*}{$Old_{Tasks}$} & $L_{ce}+L_{KD}$ & \rule[2pt]{0.9em}{0.05em}   &54.70	&33.35	&27.40	&20.65	&19.76	&17.62	&13.44	&12.05	&10.54	&23.28\\
          & $\pmb{L_{IC}}+L_{KD}$    & \rule[2pt]{0.9em}{0.05em}  &65.50	&46.00	&38.40	&31.40	&28.82	&27.48	&22.47	&20.01	&17.51	&33.07\\
          & $L_{ce}+\pmb{L_{OC}}$   & \rule[2pt]{0.9em}{0.05em}     &66.40	&47.10	&39.23	&35.00	&33.64	&32.02	&25.91	&23.76	&22.12	&36.13\\
          & $\pmb{L_{JIOC}}$  & \rule[2pt]{0.9em}{0.05em}  &65.90	&46.35	&40.27	&36.10	&33.56	&31.57	&26.83	&23.31	&22.40	&\textbf{36.25} \\
    \bottomrule
    \bottomrule
    \end{tabular}}%
   \caption{The results obtained by running with $N_{old}=1000$, using ResNet18 as the feature extractor on the CIFAR100 dataset ($L_{ce}+L_{KD}$ is the loss function used by the ICARL algorithm.)}
   \label{tb:CIFAR100_res18}
\end{table*}%

\begin{table*}[!t]
  \centering
  \resizebox{1.\linewidth}{!}{
    \begin{tabular}{c|c|c|c|c|c|c|c|c|c|c|c|c}
    \toprule
    \toprule
    \multicolumn{2}{c}{\multirow{2}[4]{*}{ICARL}} & \multicolumn{11}{c}{Task} \\
\cmidrule{3-13}    \multicolumn{2}{c}{} & 1     & 2     & 3     & 4     & 5     & 6     & 7     & 8     & 9     & \multicolumn{1}{c}{10} & Avg \\
    \midrule
    \midrule
    \multirow{4}[2]{*}{$All_{Tasks}$} & $L_{ce}+L_{KD}$ &89.90	&75.05	&67.20	&55.50	&49.62	&45.03	&41.26	&35.62	&32.77	&30.06	&52.20\\
          & $\pmb{L_{IC}}+L_{KD}$ &89.90	&74.70	&66.57	&55.65	&51.84	&45.60	&43.76	&37.16	&34.18	&33.69	&53.31\\
          & $L_{ce}+\pmb{L_{OC}}$  &89.90	&75.85	&68.97	&58.98	&54.98	&50.02	&47.26	&41.30	&39.53	&37.21	&56.40 \\
          & $\pmb{L_{JIOC}}$  &89.90	&76.25	&67.47	&58.82	&55.54	&51.77	&48.39	&42.26	&39.01	&37.23	& \textbf{56.66} \\
    \midrule
    \multirow{4}[2]{*}{$New_{Task}$} & $L_{ce}+L_{KD}$ &89.90	&78.50	&87.80	&82.70	&90.90	&83.20	&88.20	&84.00	&87.90	&83.30	&85.64 \\
          & $\pmb{L_{IC}}+L_{KD}$   &89.90	&77.80	&88.70	&83.30	&89.30	&84.70	&89.00	&85.00	&88.20	&83.40	&\textbf{85.93}\\
          & $L_{ce}+\pmb{L_{OC}}$    &89.90	&79.20	&89.50	&82.80	&89.10	&81.60	&87.10	&81.90	&86.10	&80.30	&84.75\\
          & $\pmb{L_{JIOC}}$  &89.90	&79.00	&88.70	&82.70	&89.80	&82.30	&88.00	&81.40	&86.80	&81.50	&85.01 \\
    \midrule
    \multirow{4}[2]{*}{$Old_{Tasks}$} & $L_{ce}+L_{KD}$ & \rule[2pt]{0.9em}{0.05em}   &71.60	&56.90	&46.43	&39.30	&37.40	&33.43	&28.71	&25.88	&24.74	&40.49\\
          & $\pmb{L_{IC}}+L_{KD}$    & \rule[2pt]{0.9em}{0.05em}  &71.60	&55.50	&46.43	&42.48	&37.78	&36.22	&30.33	&27.42	&28.17	&41.77\\
          & $L_{ce}+\pmb{L_{OC}}$   & \rule[2pt]{0.9em}{0.05em}     &72.50	&58.70	&51.03	&46.45	&43.70	&40.62	&35.50	&33.71	&32.42	&46.07\\
          & $\pmb{L_{JIOC}}$  & \rule[2pt]{0.9em}{0.05em}     
          &73.50	&56.85	&50.87	&46.98	&45.66	&41.78	&36.67	&33.04	&32.31	&\textbf{46.41} \\
    \bottomrule
    \bottomrule
    \end{tabular}}%
    \caption{The results obtained by running with $N_{old}=1000$, using ResNet32 as the feature extractor on the CIFAR100 dataset ($L_{ce}+L_{KD}$ is the loss function used by the ICARL algorithm.)}
  \label{tb:CIFAR100_res32}
\end{table*}%

\end{document}